\documentclass[11pt]{article}

\PassOptionsToPackage{numbers,sort&compress}{natbib}
\usepackage[final]{acl}

\usepackage{times}
\usepackage{latexsym}
\usepackage{setspace}

\usepackage[T1]{fontenc}

\usepackage[utf8]{inputenc}

\usepackage{microtype}

\usepackage{inconsolata}

\usepackage{graphicx}

\usepackage{tcolorbox}
\tcbuselibrary{minted,skins,breakable}

\newtcblisting{pybox}{
  listing engine=minted,
  minted language=python,
  minted options={
    breaklines,
    fontsize=\small,
    linenos,
    numbersep=6pt,
    autogobble,
    escapeinside=||
  },
  colback=gray!3,
  colframe=gray!40,
  arc=2pt, outer arc=2pt, boxrule=0.6pt,
  left=6pt,right=6pt,top=6pt,bottom=6pt,
  breakable,
  listing only   
}

\usepackage{tcolorbox}   
\tcbuselibrary{skins}    
\tcbuselibrary{breakable}
\usepackage{xcolor}      
\usepackage{tikz}        
\usepackage{varwidth}    
\usepackage{float} 

\usepackage[utf8]{inputenc}   
\usepackage[T1]{fontenc}      
\usepackage{times}             
\usepackage{inconsolata}       
\usepackage{microtype}         

\usepackage{amsmath, amssymb, amsfonts}
\usepackage{nicefrac}          

\usepackage{xcolor}            
\usepackage[table]{xcolor}     
\usepackage{colortbl}          
\usepackage{booktabs}          
\usepackage{multirow}          
\usepackage{makecell}          
\usepackage{array}             
\usepackage{adjustbox}         

\usepackage{graphicx}          
\usepackage{tikz}              
\usepackage{subcaption}        
\usepackage{float,wrapfig}     

\usepackage{hyperref}          
\usepackage{url}               

\usepackage{algorithm}
\usepackage{algorithmic}

\usepackage{pifont}            
\usepackage{xspace}            
\usepackage{amsthm}            
\usepackage{soul}              
\usepackage[normalem]{ulem}              
\usepackage{caption}           
\usepackage{balance}           
\setcitestyle{square,numbers}

\usepackage{pifont} 
\usepackage{xcolor} 

\definecolor{cgreen}{RGB}{34,139,34} 
\definecolor{cred}{RGB}{178, 34, 34}

\newcommand{\cmark}{\textcolor{cgreen}{\ding{51}}} 
\newcommand{\xmark}{\textcolor{cred}{\ding{55}}}   

\newcommand{\tightmidrule}{\specialrule{\lightrulewidth}{0pt}{3pt}}
\newcommand{\tightbottomrule}{\specialrule{\heavyrulewidth}{0pt}{0pt}}

\title{VideoStir: Understanding Long Videos via Spatio-Temporally Structured and Intent-Aware RAG}

\author{Honghao Fu$^{1}$, Miao Xu$^{1}$, Yiwei Wang$^{2}$, 
        \textbf{Dailing Zhang$^{3}$, Jun Liu$^{4}$, Yujun Cai$^{1}$\thanks{Corresponding Author.}} \\
        $^1$University of Queensland, \quad$^2$University of California, Merced \\
        $^3$Institute of Automation, CAS, \quad$^4$Lancaster University \\
        \normalsize \texttt{\{honghao.fu, miao.xu, yujun.cai\}@uq.edu.au}, \quad\texttt{yiweiwang2@ucmerced.edu} \\
        \normalsize \texttt{zhangdailing2023@ia.ac.cn}, \quad\texttt{j.liu81@lancaster.ac.uk} \\
        }

\begin{document}
\maketitle
\begin{abstract}

Scaling multimodal large language models (MLLMs) to long videos is constrained by limited context windows. While retrieval-augmented generation (RAG) is a promising remedy by organizing query-relevant visual evidence into a compact context, most existing methods (i) flatten videos into independent segments, breaking their inherent spatio-temporal structure, and (ii) depend on explicit semantic matching, which can miss cues that are implicitly relevant to the query’s intent. To overcome these limitations, we propose VideoStir, a structured and intent-aware long-video RAG framework. It firstly structures a video as a spatio-temporal graph at clip level, and then performs multi-hop retrieval to aggregate evidence across distant yet contextually related events. Furthermore, it introduces an MLLM-backed intent-relevance scorer that retrieves frames based on their alignment with the query’s reasoning intent. To support this capability, we curate IR-600K, a large-scale dataset tailored for learning frame–query intent alignment. Experiments show that VideoStir is competitive with state-of-the-art baselines without relying on auxiliary information, highlighting the promise of shifting long-video RAG from flattened semantic matching to structured, intent-aware reasoning. Codes and checkpoints are available at \url{https://github.com/RomGai/VideoStir}.
\end{abstract}

\section{Introduction}
\label{sec:intro}

\begin{figure}[t]
    \centering
    \includegraphics[width=\linewidth]{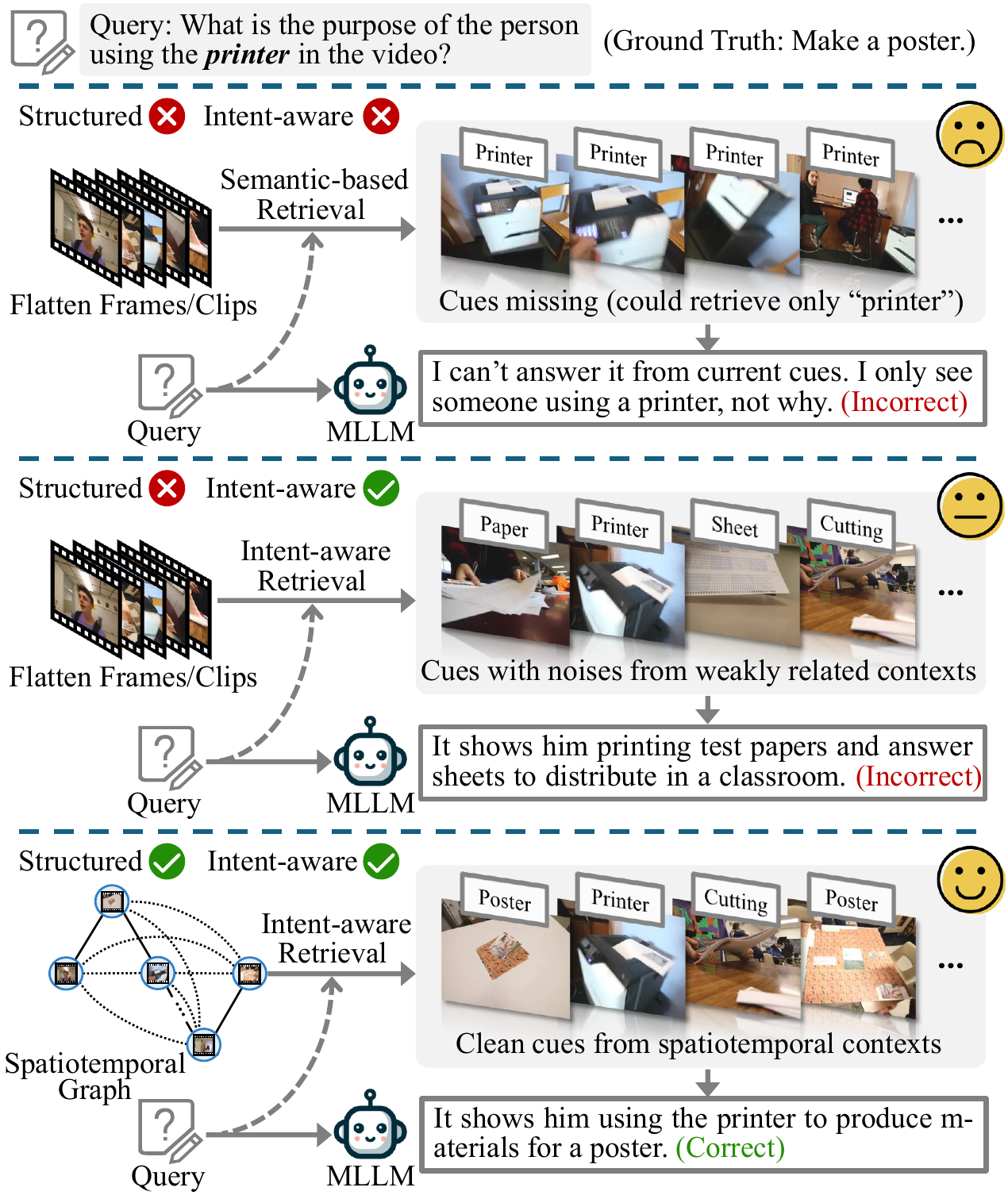}
    \caption{\small 
Paradigm shift from flat semantic matching to structured, intent-aware long-video RAG. \textbf{(Top)} Semantic-based retrieval relies on explicit semantic overlap, often missing cues that are only implicitly relevant to the query intent. \textbf{(Middle)} Intent-aware retrieval utilizes MLLM's reasoning capability to identify cues that may be relevant to the query intent (detailed in Sec.~\ref{sec:method_int}); however, it can be distracted by noisy cues when operating over flattened contexts scattered across long temporal ranges. \textbf{(Bottom)} VideoStir structures the video as a spatio-temporal graph to retrieve cues from coherent contexts, enabling more reliable evidence aggregation. We further discuss the benefits of structured retrieval in Fig.~\ref{fig:extended_case}.}
    \label{fig:moti}
    \vspace{-0.15in}
\end{figure}

Video understanding has emerged as a core frontier in multimodal intelligence~\cite{li2025videochat}. While recent Multimodal Large Language Models (MLLMs) have achieved {remarkable success in interpreting short clips, extending this capability to long videos remains a formidable challenge~\cite{Video-RAG}. The primary difficulty lies not merely in the video's length, but in its information density and structural complexity~\cite{vgent}. Unlike short clips that typically encapsulate a single event, long videos consist of interleaved narratives where critical cues could be sparsely scattered across the timeline. To handle the extensive contexts, state-of-the-art (SOTA) models often resort to expanding context windows to support uniform sampling at second level~\cite{comanici2025gemini25}.
However, such approaches face a trade-off: they could either miss fine-grained details due to temporally sparse sampling or struggle with accurate reasoning when overwhelmed by redundant visual noise.}

To address the contextual intractability of processing long videos, Retrieval-Augmented Generation (RAG) has emerged as a promising solution by retrieving a small subset of segments to fit within limited context windows~\cite{exvideorag}. However, current long-video RAG methods typically flatten a video into a set of independent segments for query-based retrieval~\cite{AKS,FOCUS}, and rank candidates primarily by embedding similarity computed with contrastive vision–language models (e.g., CLIP)~\cite{clip,Video-RAG,TVRAG}. Yet this flattened semantic matching paradigm leaves two critical gaps that hinder downstream reasoning.

The first gap is \textbf{spatio-temporal structure decoupling}. Flattening a video into ndependent segments disrupts its intrinsic spatio-temporal structure, decoupling contexts that should remain linked. Because text queries may lack the fine-grained cues needed to reconnect dispersed evidence, direct matching often struggles to retrieve contexts that are spatio-temporally relevant yet share little explicit semantic overlap with the query, limiting downstream models’ ability to form coherent reasoning chains~\cite{vgent,liao2024videoinsta}. The second gap is \textbf{insufficient intent modeling}. Similarity computed from contrastive embeddings is optimized for vision–language alignment and tends to capture ``what looks similar'' rather than ``why it matters'' for addressing the query’s reasoning intent~\cite{pan2024i3}. As shown in Figure~\ref{fig:moti}, the query ``For what \textit{purpose} does the recorder use the \textit{printer}?'' leads a semantic retriever to select frames showing the ``\textit{printer}'' due to high semantic overlap, whereas the actual ``\textit{purpose}'' is revealed in a different event (e.g., making a poster), which semantic overlap alone fails to surface.

In light of these gaps, we advocate two shifts in long-video RAG paradigms: \textit{(i) From Flattened to Structured}: moving beyond isolated segment retrieval to reconstruct the intrinsic spatio-temporal topology of the video; and \textit{(ii) From Semantic to Intent}: going beyond surface semantic to model the alignment between query's intent and visual cues.

{To realize these shifts, we propose \textbf{VideoStir}, a framework designed to advance long-video RAG from flattened semantic matching to \textbf{ST}ructured, \textbf{I}ntent-aware \textbf{R}easoning.} Inspired by human episodic memory where recall proceeds coarse-to-fine (locating relevant episodes before examining details), VideoStir operates in two phases, from clips to frames. First, a coarse clip-retrieval phase addresses structure decoupling by representing the video as a spatio-temporal graph, whose nodes are semantically coherent short clips segmented by an event boundary detector. Temporal edges preserve narrative continuity, while spatial edges bridge distant yet semantically related clips via inter-clip similarity. Traversing this topology enables multi-hop retrieval from query-matched anchors to their spatio-temporal neighbors, leveraging intrinsic correlations among semantically dense clips to aggregate dependencies and mitigate the semantic sparsity of direct query matching. Second, a fine-grained frame-retrieval phase addresses insufficient intent modeling with an \textbf{I}ntent-\textbf{R}elevance scorer trained on our curated \textbf{IR-600K} dataset. Rather than considering semantic similarity alone, this lightweight MLLM-based scorer estimates frame–intent alignment, ensuring retrieved frames are not only semantically relevant but also pivotal for downstream reasoning.

Consequently, VideoStir provides the downstream MLLM with spatio-temporally context-coherent and intent-aligned visual cues, enabling more accurate long-video understanding. The scorer and curated IR-600K dataset further provide a reusable foundation for future research on intent-oriented long-video RAG systems. Our contributions are summarized as follows:

\begin{itemize}
\setlength{\itemsep}{2pt}
\setlength{\parsep}{2pt}
\setlength{\parskip}{2pt}
    \item We propose VideoStir, a novel long-video RAG framework that leverages structured retrieval to reconstruct video's spatio-temporal topology and aggregate contextually coherent evidence, which overcomes the spatio-temporal decoupling of flattened retrieval.
    \item We introduce an intent-relevance scorer and IR-600K, a large-scale dataset modeling the relevance between frames and query intent. This shifts retrieval from semantic matching to intent awareness, establishing a foundation for intent-oriented long-video RAG.
    \item Extensive experiments show that VideoStir is competitive with SOTA baselines without relying on auxiliary information, highlighting the promise of the paradigm shift from flattened semantic matching to structured and intent-aligned reasoning in long-video RAG.

\end{itemize}

    

    


\nocite{liu2025hacsurv,zhang2024defending,zhang2025tuning,zhang2025improving,mei2026gateddifferentiableworkingmemory}
\section{Related Works}
\label{sec:related_works}
\subsection{Video-Language Models}
Long-video understanding has attracted increasing attention in multimodal intelligence. Recent general-purpose MLLMs such as Gemini~\cite{gemini} and Long-VITA~\cite{longvita}, along with specialized video models including Video-XL~\cite{videoxl} and LongVLM~\cite{longvlm}, extend visual context to enable hour-scale video reasoning via uniform sampling. However, this paradigm faces a major bottleneck: limited context windows necessitate sparse sampling across the timeline, which can yield semantically redundant frames while simultaneously risking the loss of fleeting, query-critical cues. In parallel, contrastive video–language models~\cite{clip4clip} (e.g., Video-CLIP~\cite{videoclip}, X-CLIP~\cite{xclip}, and PE~\cite{perceptionenc}) extend CLIP-style contrastive embeddings to video, enabling query-conditioned retrieval of salient clips and frames. Nevertheless, contrastive objectives primarily optimize for semantic similarity and may miss cues that are implicitly relevant to the query intent but lack an explicit semantic match.

\begin{figure*}[t]
    \centering
    \includegraphics[width=\linewidth]{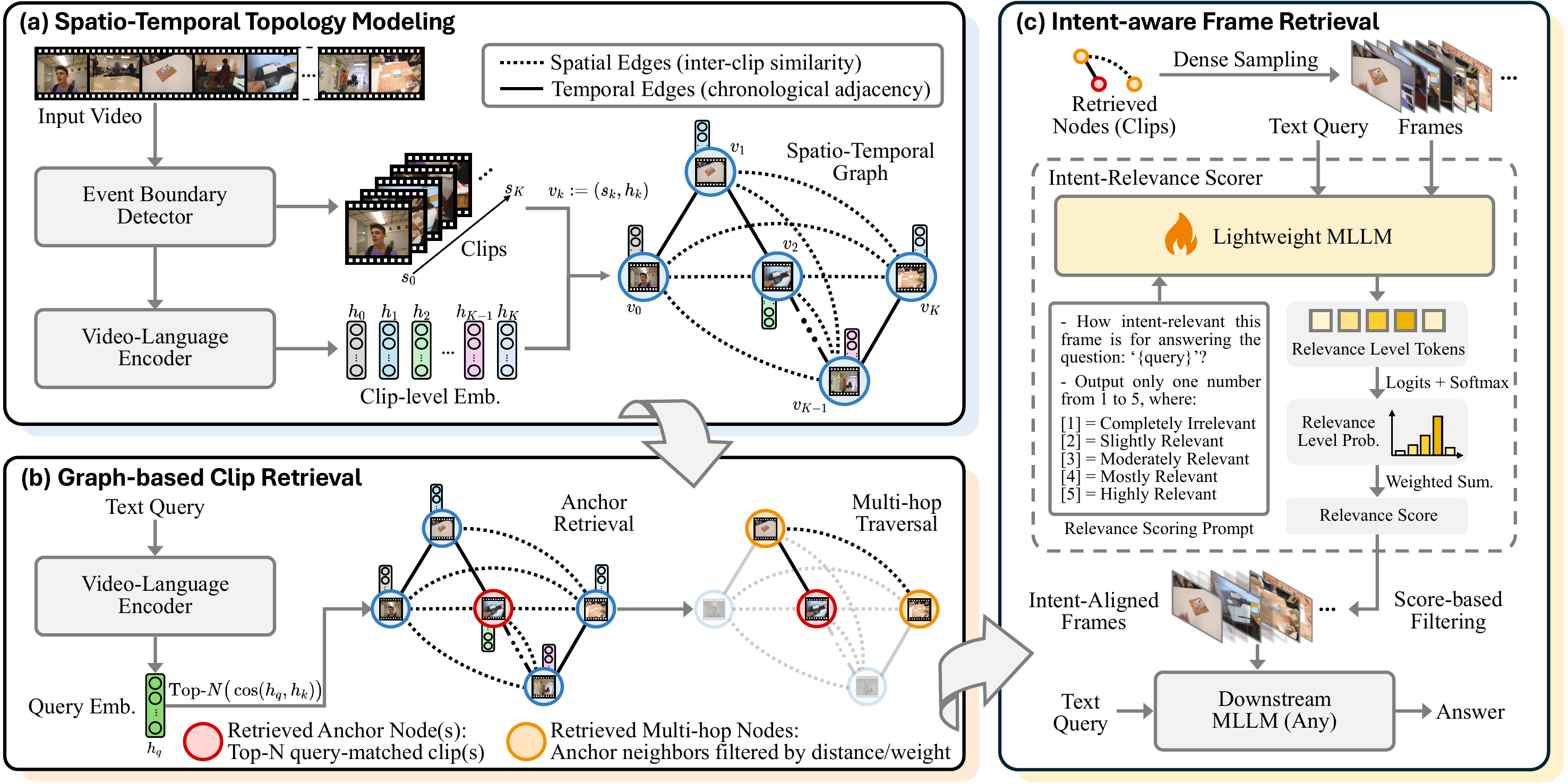}
    \caption{\small {\textbf{Overview of VideoStir.}} VideoStir achieves long-video understanding via spatio-temporal structuring and two-stage retrieval. \textbf{(a) Spatio-Temporal Topology Modeling.} An event boundary detector segments input video into semantically coherent clips. Their embeddings are used to construct a spatio-temporal graph with temporal edges linking adjacent clips and spatial edges weighted by inter-clip similarity. \textbf{(b) Graph-based Clip Retrieval}. Input query is embedded in the video-language space to retrieve top-$N$ matched anchor nodes; multi-hop traversal then expands along spatio-temporal links under hop-distance and edge-weight constraints to aggregate contextual clips. \textbf{(c) Intent-aware Frame Retrieval.} Frames densely sampled from the retrieved clips are assessed by an intent-relevance scorer (LoRA-tuned on IR-600K), which outputs a probability distribution over discrete relevance levels. These distributions are then aggregated via probability-weighted expectation to produce a continuous relevance score; high-scoring, intent-aligned frames are retained and fed to a downstream MLLM to produce the final answer.}
    \label{fig:fw}
\end{figure*}

\nocite{gu2025optical,xiong2025unveiling,wu2025refineshot,gu2025acl,li2024think,ge2025focusingcontrastiveattentionenhancing,fu2024dp,gu2025hdtcnet,zhang2025tokenswap}

\subsection{Agentic Long-Video Understanding}
To alleviate the contextual intractability inherent in processing long-duration videos, a growing body of recent work has shifted towards leveraging external systems. These systems are designed to reconstruct and distill extensive long-range context into compact, semantically rich cues tailored for downstream MLLMs. A prominent trajectory within this domain involves the exploration of agentic frameworks~\cite{mmvid,omagent}, which employ (M)LLMs as decision-making policies equipped with advanced modules for captioning, multi-step reasoning, and external memory~\cite{VideoMem}. By simulating human-like curation processes~\cite{vamos,morevqa}, these agents dynamically filter irrelevant data. For example, DrVideo~\cite{drvideo} synthesizes disjoint video frames into coherent long-form textual narratives to facilitate precise query answering, while Vgent~\cite{vgent} abstracts complex entity interactions into structured symbolic representations to bolster reasoning. However, despite their demonstrated efficacy in enhancing understanding, deploying such sophisticated MLLM-based policies incurs substantial inference overhead.
\subsection{RAG for Long-Video Understanding}
{To reduce the procedural inference overhead of agentic workflows, another line of work explores long-video retrieval-augmented generation (RAG) frameworks~\cite{exvideorag,xue2025omni,zeng2025scenerag}, which externally select cues to fit limited context windows~\cite{videorag,TVRAG,FOCUS}. For example, Video-RAG~\cite{Video-RAG} retrieves keyframes via semantic similarity, and enriches entity information with external tools like optical character recognition and object detectors. E-VRAG~\cite{evrag} introduces a retrieval-reranking pipeline, leveraging Qwen3-Reranker-style binary judgments for fine-grained selection. while AKS~\cite{AKS} selects keyframes by jointly optimizing query-frame semantic similarity and temporal uniformity. However, existing methods often treat videos as independent segments, decoupling intrinsic spatio-temporal structure and limiting coherent reasoning. Moreover, current paradigms favor explicit semantic matching over intent-level alignment, risking the omission of cues implicitly relevant to the query intent.}


\nocite{fu2025contextnav,liu2025correlationcausationmaxpoolingbasedmultiinstance,ge2025framemind,gu2025sfir,gu2025efficient,fu2023sgcn,mei2025a1,li2024drs,li2025texture,chen2025haif}
\section{Method}
\subsection{Overview}
We propose VideoStir, a long-video RAG framework that explicitly models video spatio-temporal topology and query intent to retrieve pivotal evidence, thereby advancing downstream MLLMs’ long-video understanding. 

As illustrated in Figure~\ref{fig:fw}, VideoStir comprises three phases: (1) Spatio-temporal topology modeling. It converts the input video into a structured graph. An event boundary detector partitions the video into clip-level nodes connected by temporal edges (chronological adjacency) and spatial edges (inter-clip semantic similarity). This design reconstructs the topology of the video’s spatio-temporal context. (2) Graph-based clip retrieval. It first identifies anchor clips semantically aligned with the query, then expands the context via multi-hop traversal on the spatio-temporal graph. Consequently, the retrieved evidence includes not only query-matched moments but also their spatio-temporally relevant neighbors that are critical for reasoning. (3) Intent-aware frame retrieval. It refines evidence at the frame level. Since semantic similarity often fails to capture the query’s underlying intent, we introduce an intent-relevance scorer. Specifically, we leverage a lightweight MLLM fine-tuned on our curated IR-600K dataset to assign each frame a relevance score, computed as the weighted expectation over predicted relevance levels. It mitigates the loss of implicit cues inherent to explicit semantic matching, yielding an intent-aligned evidence set for downstream MLLMs.

\subsection{Spatio-Temporal Topology Modeling}

To structurally encode a long video’s spatio-temporal topology, we model it as a graph $\mathcal{G}=(\mathcal{V},\mathcal{E})$, where each node $v_k \in \mathcal{V}$ corresponds to a clip $s_k \subset S$. These clips are obtained via an event boundary detector, which preprocesses the long video into semantically coherent segments by identifying event transitions through semantic change-point detection. This process is implemented by applying PELT~\cite{PELT} to the frame embeddings for adaptive segmentation. For each clip, we compute an embedding $h_k = E_{\mathrm{vl}}(s_k)$ with a video–language encoder $E_{\mathrm{vl}}$. The edge set $\mathcal{E}=\{\mathcal{E}_{\mathrm{Temp}}, \mathcal{E}_{\mathrm{Sp}}\}$ consists of temporal edges $\mathcal{E}_{\mathrm{Temp}}=\{(v_k,v_{k+1})\mid w_{k,k+1}=1\}$ and spatial edges $\mathcal{E}_{\mathrm{Sp}}=\{(v_i,v_j)\mid w_{i,j}=\cos(h_i,h_j)\}$, where $w$ denotes the edge weight. {Here, we refer to the embedding space as the ``spatial'' dimension of the topology. Temporal edges connect chronologically adjacent clips, preserving the video’s temporal backbone. Spatial edges are weighted by the similarity between clip embeddings, capturing inter-clip correlations when different moments share similar video-level features. By jointly modeling these edges, $G$ induces a spatio-temporal topology over clips: temporally adjacent events are linked along timeline, while semantically coherent yet temporally distant events are bridged via spatial edges. This graph re-entangles contextual relations that may be lost in a flattened representation and provides the backbone for multi-hop retrieval.}

\subsection{Graph-based Clip Retrieval}

Given a query $q$ and a spatio-temporal graph $\mathcal{G}$, we aim to coarsely retrieve clips that are semantically relevant to $q$ and topologically supported by inter-clip correlations. {This design addresses semantically sparse query–clip matching: a query may refer to only a small facet of an event (e.g., the main subject), whereas long videos often contain recurring scenes/actions/entities and temporally adjacent evidence that constitute critical spatio-temporal contexts for the underlying event, which can be missed by simply computing semantic similarity between the query and individual moments. As shown in Figure~\ref{fig:extended_case}, with our spatio-temporal graph, these contexts can be recovered via inter-clip similarity and chronological adjacency encoded by spatio-temporal edges. 

Specifically, we first compute the query embedding $h_q = E_{\text{vl}}(q)$ using the same video-language encoder as for clips, placing queries and clip nodes in a shared embedding space. We then measure query-node similarity via $\cos(h_q, h_k)$, and select the top-$N$ similar nodes as the anchor set $\mathcal{V}_{\text{anc}}$. While these anchors are matched to the query, they may not fully cover their implicitly contextual dependencies. To recover broader context, we expand $\mathcal{V}_{\text{anc}}$ along spatio-temporal links in $\mathcal{G}$, traversing edges according to their weights and collecting nodes within $L$ hops of the anchors:
\begin{equation}
\resizebox{0.75\linewidth}{!}{$\displaystyle
\mathcal{V}_\text{hop}
= \{\, v_j \mid d(v_j, \mathcal{V}_{\text{anc}}) \le L,\; w_{ij} \ge \eta \,\},
$}
\end{equation}
where $d(v_j, \mathcal{V}_{\text{anc}})$ is the shortest hop distance from $v_j$ to any anchor in $\mathcal{V}_{\text{anc}}$, and threshold $\eta$ filters out weak connections. {In this way, multi-hop traversal yields a coarse spatio-temporal neighborhood around the event hinted by the query, providing contextual candidates for subsequent fine-grained intent-level filtering.}

\subsection{Intent-aware Frame Retrieval}
\label{sec:method_int}
\noindent\textbf{Intent Relevance Reasoning and Retrieval.} While multi-hop retrieval yields a set of contextually related clips, the candidates may include redundant or merely co-occurring frames that are semantically relevant but misaligned with the query’s underlying intent. We therefore introduce an intent-relevance scorer $R_\theta$, which filters retrieved clips at frame level using the intent-reasoning capability of a lightweight MLLM backbone $\pi_{\theta}$. Specifically, $R_\theta$ estimates frame-query intent relevance, separating frames that genuinely support the query’s intent from those with only superficial similarity.

Given the retrieved node set $\mathcal{V}_{\text{hop}} = \{ v_k \}$,  where each node represents a video clip, we expand them into a dense frame pool: $\mathcal{F}_{\text{ret}} = 
\bigcup_{v_k \in \mathcal{V}_{\text{hop}}} 
\{ x_t \mid x_t \in v_k \}$. For a query–frame pair $(q, x_t)$, the scorer outputs a softmax-normalized probability distribution over discrete intent-relevance level tokens $\ell \in \{1,2,3,4,5\}$:
\begin{equation}
\resizebox{\linewidth}{!}{$\displaystyle
P_{\theta}(\ell \mid q, x_t, \mathcal{P}_{\text{intent}})
= \frac{\exp(\pi_{\theta}(\ell \mid q, x_t, \mathcal{P}_{\text{intent}}))}
{\sum_{k=1}^{5}\exp(\pi_{\theta}(k \mid q, x_t, \mathcal{P}_{\text{intent}}))},
$}
\end{equation}
where $\pi_{\theta}(\cdot)$ denotes the logits produced by the MLLM backbone of $R_{\theta}$, and $\mathcal{P}_{\text{intent}}$ (detailed in Appendix~\ref{appendix:prompt}) denotes the prompt that guides the MLLM backbone $\pi_\theta$ to reason and score intent relevance. The score of frame $x_t$ is computed as the weighted expectation over relevance level tokens:
\begin{equation}
\resizebox{\linewidth}{!}{$r_t = R_{\theta}(q, x_t,\mathcal{P}_{\text{intent}}) = \sum_{\ell=1}^{5} \ell \cdot P_{\theta}(\ell \mid q, x_t,\mathcal{P}_{\text{intent}}).
$}
\end{equation}
This converts the model’s internal reasoning into a continuous relevance score, distinguishing frames that provide intent-level support for the query from those that are merely semantically similar. After obtaining scores for $\mathcal{F}_{\text{ret}}$, 
we retain frames exceeding a relevance threshold $\kappa_s$. 
The resulting set $\mathcal{F}_{\text{intent}} =
\{\, x_t \mid r_t > \kappa_s \,\}$ provides an intent-aligned visual context for downstream reasoning.

\noindent\textbf{Distillation-based Scorer Training.} To strengthen the scorer’s intent-relevance assessment, we adopt a distillation strategy to fine-tune its MLLM backbone. Specifically, given a query-frame pair $(q, x_t)$ and the intent-relevance scoring prompt $\mathcal{P}_\text{intent}$, a powerful teacher MLLM $\mathcal{T}$ is asked to rate the intent relevance between $q$ and $x_t$ on the discrete level (1-5). The resulting pseudo-label $y_t = \mathcal{T}(q, x_t, \mathcal{P}_\text{intent})$ serves as supervision for the student scorer $R_{\theta}$. During fine-tuning, we inject low-rank adaptation (LoRA) parameters~\cite{lora} into the MLLM backbone of $R_{\theta}$, enabling efficient adaptation without substantially disrupting its pretrained priors. Given the student prediction $P_{\theta}(\ell \mid q, x_t, \mathcal{P}_\text{intent})$, the training objective minimizes the cross-entropy between the student-predicted distribution and the teacher-provided discrete label:
\begin{equation}
\resizebox{0.85\linewidth}{!}{$\displaystyle
\mathcal{L}_{\mathrm{CE}} = 
- \sum_{\ell=1}^{5} 
\mathbf{1}[\ell = y_t] \log P_{\theta}(\ell \mid q, x_t, \mathcal{P}_{\text{intent}}).
$}
\end{equation}
The LoRA parameters are optimized via gradient descent, enabling the student to inherit the teacher’s relevance-assessment patterns and yield more accurate relevance scores, while retaining the computational efficiency of its lightweight backbone.

\nocite{gu2023orsi,ge2025innate,gu2024mixed,mei2025surveycontextengineeringlarge,wu2025dimo,mei2024hiddenguard,li2024vulnerability,chen2025tokensnodessemanticguidedmotion,mei2024not}
\section{Experiment}

\begin{table*}[t]
\centering
\small
\caption{\small {Comparison with long-video RAG baselines. ``Native Input'' indicates whether the model processes video directly without relying on external auxiliary text or tools. Bold and underlined numbers indicate the best and second-best accuracies, respectively. ‘–’ indicates not applicable.}}
\resizebox{\linewidth}{!}{
\begin{tabular}{lc cc cc cc}
\toprule
\multirow{2}{*}{\textbf{Model}} & \multirow{2}{*}{\textbf{Native Input}} &
\multicolumn{2}{c}{\textbf{LV-Bench (Val)}} &
\multicolumn{2}{c}{\textbf{MLVU}} &
\multicolumn{2}{c}{\textbf{Video-MME-Long (w/o Sub)}} \\
\cmidrule(lr){3-4} \cmidrule(lr){5-6} \cmidrule(lr){7-8}
 & & Overall $\uparrow$ & Gain (\%) $\uparrow$ &
     Overall $\uparrow$ & Gain (\%) $\uparrow$ &
     Overall $\uparrow$ & Gain (\%) $\uparrow$ \\
\midrule
LLaVA-Video (7B) & \cmark & 56.6 & - & 70.8 & - & - & - \\
+ Video-RAG & \xmark & 58.7 & 3.7 & \underline{72.4} & \underline{2.3} & - & - \\
+ TV-RAG & \xmark & 58.8 & 2.1& 72.1 & 1.8&- & - \\
\rowcolor[rgb]{0.80,0.90,0.95} \textbf{+ VideoStir (Ours)} & \cmark & \textbf{60.3} & \textbf{6.5} & \textbf{73.1}& \textbf{3.2} & - & - \\ 
\tightmidrule
mPLUG-Owl3 (8B) & \cmark & 52.1 & - & 63.7 & - &50.1 & - \\
+ Video-RAG & \xmark & 54.5   & 4.6 & \underline{64.4} & \underline{1.1}& 50.6 & 1.0 \\
+ TV-RAG & \xmark & 54.8 & 5.2& 64.0 &0.5 & \underline{50.9}& \underline{1.6} \\
\rowcolor[rgb]{0.80,0.90,0.95} \textbf{+ VideoStir (Ours)} & \cmark & \textbf{55.1} & \textbf{5.8} & \textbf{64.6} & \textbf{1.4} & \textbf{51.0} &\textbf{1.8}  \\
\tightmidrule
Aria (25B) & \cmark & 64.2 & - & 70.6 & - & 58.8 & - \\
+ Video-RAG & \xmark & \textbf{66.4} & \textbf{3.4} & \textbf{72.1} & \textbf{2.1} & \textbf{59.6}& \textbf{1.4} \\
+ TV-RAG & \xmark & 65.5 & 2.0& 71.2& 0.8&59.0& 0.3\\
\rowcolor[rgb]{0.80,0.90,0.95} \textbf{+ VideoStir (Ours)} & \cmark & \underline{65.8} & \underline{2.5} &  \underline{71.7} & \underline{1.6} & \underline{59.4} & \underline{1.0} \\
\tightmidrule
InternVL-1.5 (26B) & \cmark & 51.2 & - & 50.4 & - & 45.6 & - \\
+ Video-RAG& \xmark & 52.2 & 2.0 &\textbf{51.8} & \textbf{2.8} & \underline{46.7} & \underline{2.4} \\
+ TV-RAG & \xmark & \underline{52.4} & \underline{2.3}& 51.0 &1.2 & 46.3& 1.5\\
\rowcolor[rgb]{0.80,0.90,0.95} \textbf{+ VideoStir (Ours)} & \cmark & \textbf{52.7} & \textbf{2.9} & \underline{51.6}&\underline{2.4} &\textbf{46.9} & \textbf{2.9} \\ 
\tightmidrule
LLaVA-Video (72B) & \cmark & 61.9 & - & 73.1 & - &61.5 & - \\
+ Video-RAG & \xmark & \underline{65.4} & \underline{5.7} & \underline{73.8} & \underline{1.0} & \textbf{62.3} & \textbf{1.3} \\
+ TV-RAG & \xmark & 64.6 & 4.3 &  73.4& 0.4& 61.8 & 0.5 \\
\rowcolor[rgb]{0.80,0.90,0.95} \textbf{+ VideoStir (Ours)} & \cmark & \textbf{66.0} &\textbf{6.6}&  \textbf{74.1}  & \textbf{1.4} & \underline{62.1} & \underline{1.0} \\ 
\tightbottomrule
\end{tabular}
}
\label{tab:comp1}
\end{table*}

\begin{table}[t]
\centering
\small
\caption{\small {System-level comparison on EgoSchema's test set with SOTA methods built upon fixed (M)LLMs. These approaches can be categorized into two groups: one employs video-to-text tools (e.g., captioners or object detectors) to provide auxiliary textual inputs for downstream (M)LLMs, while the other leverages the native multimodal understanding capability (Native Input) of MLLMs without auxiliary text. Bold and underlined numbers indicate the best and second-best accuracies, respectively. ‘–’ indicates not applicable.}}
\resizebox{\linewidth}{!}{
\begin{tabular}{l l c}
\toprule
(M)LLM Backbone & Method & Overall$\uparrow$ \\
\specialrule{\lightrulewidth}{0pt}{0pt}
\rowcolor[HTML]{FFF7F0}
\multicolumn{3}{c}{\rule[-0.9ex]{0pt}{3.3ex}\textbf{\textit{Methods with Video-to-Text Tools}}}\\
\specialrule{\lightrulewidth}{0pt}{0.8ex}
\multirow{2}{*}{GPT-3.5} & LLoVi~\cite{llovi} & 57.6 \\
                       & DrVideo~\cite{drvideo}        & 62.6 \\
\midrule
\multirow{4}{*}{GPT-4} & VideoAgent~\cite{videoagent} & 60.2 \\
                       & LLoVi~\cite{llovi}          & 61.2 \\
                       & VidAgent~\cite{vidagent}  & 62.8 \\
                       & VideoTree~\cite{videotree}  & 66.2\\
                       & DrVideo~\cite{drvideo}               & 66.4 \\
\midrule
Qwen2-VL-7B     & Vgent~\cite{vgent}    &  \textbf{68.0}\\
\specialrule{\lightrulewidth}{0pt}{0pt}
\rowcolor[HTML]{FFF7F0}
\multicolumn{3}{c}{\rule[-0.9ex]{0pt}{3.3ex}\textbf{\textit{Methods with Native Input}}}\\
\specialrule{\lightrulewidth}{0pt}{0.8ex}
\multirow{2}{*}{Qwen2-VL-7B} & IG-VLM~\cite{igvlm} & 66.2 \\
                            & VideoStir (Ours)         & \underline{67.2}\\
\bottomrule
\end{tabular}
}
\label{tab:comp2}
\end{table}


\subsection{Dataset and Implementation}
\noindent\textbf{Dataset.} We curate IR-600K, a  dataset designed to enhance small-scale MLLMs’ ability to judge intent relevance between video frames and queries, supporting intent-aware long-video RAG. To the best of our knowledge, IR-600K is the first dataset targeting intent-level frame-query alignment, i.e., whether a frame provides evidence that satisfies the query’s underlying information need. This differs from prior datasets such as TVSum~\cite{tvs} and QVHighlights~\cite{qah}, which primarily evaluate the semantic saliency of frames with respect to descriptive statements. 

Specifically, we sample 4,678 videos from three representative video QA datasets (ActivityNet-QA~\cite{A-QA}, NExT-QA~\cite{N-QA}, and STAR~\cite{S-QA}).
For each video, we uniformly extract frames at 3~fps to form query--frame pairs. We then distill supervision from a powerful MLLM: Qwen2.5-VL-72B-Instruct~\cite{bai2025qwen2} serves as the teacher to produce frame-level intent-relevance labels. Guided by a carefully designed system prompt specifying explicit intent-relevance criteria, the teacher assesses intent-level alignment between each frame and the query, yielding intent-relevance scores. IR-600K comprises 605,676 training samples and 21,791 validation samples. Further details on the intent-relevance scoring prompt, dataset statistics and annotation format are provided in Appendix~\ref{appendix:dataset} and~\ref{appendix:prompt}. 

\noindent\textbf{Implementation.} We adopt Qwen2.5-VL-3B-Instruct~\cite{bai2025qwen2} as the backbone of the intent-relevance scorer and fine-tune it with LoRA adapters with a rank of 16, a scaling factor of 32, and a dropout of 0.05. We optimize the LoRA using AdamW with a learning rate of $5\times10^{-5}$ and a weight decay of 0.05, together with a cosine learning-rate schedule and a warm-up ratio of 0.05. Training is performed for 1 epoch with a batch size of 128 under bf16 precision. For the event boundary detector, we use the vision tower from the Qwen2.5-VL series as the embedding backbone. For spatio-temporal graph construction and multi-hop retrieval, we use the Perception Encoder~\cite{perceptionenc} as the video-language encoder $E_{\mathrm{vl}}$ and set $N=3$, $L=2$, $\eta=0.4$, and $\kappa_s=3.25$; an ablation study on these retrieval hyperparameters is provided in Appendix~\ref{appendix:hyperparam}. All training runs and comparative experiments are conducted on 8$\times$A100 GPUs, while the remaining experiments are conducted on a single A100 GPU. Unless otherwise specified, results are reported as the median of three trials, and ablations are conducted on LLaVA-Video (7B) using LongVideoBench~\cite{lvb}.

\subsection{Comparison Against Other Methods} 
Table~\ref{tab:comp1} compares VideoStir with the native MLLM and state-of-the-art (SOTA) long-video RAG baselines on LongVideoBench (LV-Bench)~\cite{lvb}, MLVU~\cite{mlvu}, and Video-MME-Long~\cite{videomme}, following the same experimental setup as Video-RAG~\cite{Video-RAG} and TV-RAG~\cite{TVRAG}. Baseline MLLMs include GPT-4o~\cite{hurst2024gpt}, LLaVA-Video (7B/72B)~\cite{zhang2024video}, mPLUG-Owl3 (8B)~\cite{ye2024mplug}, Aria (25B)~\cite{aria}, and InternVL-1.5 (26B)~\cite{chen2024far}. Results are compiled from the benchmarks’ leaderboards and the original papers, with the remaining results obtained from our reproductions based on their open-sourced codes and original settings. In most cases, VideoStir outperforms both the native model and prior long-video RAG baselines, demonstrating highly competitive performance. Notably, VideoStir preserves the native MLLM input setting by operating solely on retrieved visual evidence: it introduces no auxiliary text, tool calls, or extra modalities, while still achieving SOTA performance.  

Table~\ref{tab:comp2} compares VideoStir on EgoSchema against specialized long-video systems built atop fixed (M)LLMs~\cite{egoschema}; baseline results are taken from DrVideo~\cite{drvideo} and Vgent~\cite{vgent}. Relying solely on native multimodal inputs without external video-to-text tools (e.g., captioners or object detectors), VideoStir achieves performance comparable to these specialized systems that leverage such auxiliary tools.

\begin{table}[t]
\centering
\small
\setlength{\tabcolsep}{3pt}
\caption{\small Component-level ablations on the expanded 1k validation set. ``Overall'' denotes overall accuracy, and ``Retrieval Acc.'' measures the accuracy of successfully retrieving key cues. ``w/o Intent-relevance Scorer'' replaces the scorer with similarity-based frame retrieval; ``w/o Weighted Expectation'' replaces probability-weighted expectation with discrete scores; ``w/o Spatio-Temporal Graph'' disables structured retrieval, which bypasses the graph and directly selects the top-14 clips by PE-embedding similarity, following the best-performing setting in Table~\ref{tab:ablation_hyperparams}. Bold indicates the best results.}
\resizebox{\linewidth}{!}{
\begin{tabular}{lccc}
\toprule
\textbf{Method} & \textbf{Overall$\uparrow$} & \textbf{Retrieval Acc.$\uparrow$}\\
\midrule
w/o Intent-relevance Scorer \\
\quad + CLIP L/14                   & 52.8 & 69.2 \\
\quad + Video-CLIP                  & 56.3 & 75.4 \\
\quad + PE                          & 58.1 & 79.8 \\
w/o Weighted Expectation            & 54.2 & 71.6 \\
w/o Event-Boundary Detector         & 62.3 & 88.2 \\
w/o Spatio-Temporal Graph           & 56.4 & 74.8 \\
w/o Spatial Edges                   & 57.2 & 79.3 \\
w/o Temporal Edges                  & 59.8 & 83.4 \\
\midrule
\textbf{Full}                 & \textbf{64.5} & \textbf{92.2}  \\
\bottomrule
\end{tabular}
}
\label{tab:ab1}
\end{table}

\begin{table}[t]
\centering
\small
\setlength{\tabcolsep}{6pt}
\renewcommand{\arraystretch}{1.12}
\caption{\small 
Ablations of VideoStir's video-language embedding backbones with $N=1000$. ``Overall'' denotes overall accuracy, and ``Retrieval Acc.'' refers to the accuracy of successfully retrieving the key evidence for answering the query. The bold numbers represent the best results.} 
\resizebox{\linewidth}{!}{
\begin{tabular}{lcc}
\toprule
\textbf{Embedding Backbone} & \textbf{Overall$\uparrow$} & \textbf{Retrieval Acc.$\uparrow$} \\
\midrule
Video-CLIP~\cite{videoclip}              & 62.1 & 88.3  \\
X-CLIP~\cite{xclip}                      & 62.8 & 89.4 \\
\textbf{PE}~\cite{perceptionenc}         & \textbf{64.5} & \textbf{92.2}  \\
\bottomrule
\end{tabular}
}
\label{tab:ab2}
\end{table}

\begin{table*}[t]
\centering
\small
\caption{\small {Ablations on the intent-relevant scorer’s distillation strategies. ``IR-600K CE'' and ``LV-Bench CE'' denote the cross-entropy loss computed on the corresponding datasets. To reduce distributional bias, the student model is distilled using teacher models from the same family. ``Retrieval Acc.'' refers to the accuracy of successfully retrieving the key evidence.}} 
\resizebox{\linewidth}{!}{
\begin{tabular}{llccccc}
\toprule
\textbf{Base Model} &\textbf{Method} & \textbf{IR-600K CE $\downarrow$}& \textbf{LV-Bench CE $\downarrow$}&\textbf{Overall $\uparrow$}&\textbf{Retrieval Acc. $\uparrow$}& \textbf{Trainable Params $\downarrow$}\\
\specialrule{\lightrulewidth}{0pt}{0pt}
\rowcolor[HTML]{FFF7F0}
\multicolumn{7}{c}{\rule[-1ex]{0pt}{3.3ex}\textbf{\textit{Teacher Models}}}\\
\specialrule{\lightrulewidth}{0pt}{0.8ex}
InternVL2.5-78B&Zero-Shot   &  -&-& 64.8 & 92.6 &  - \\
Qwen2.5-VL-72B&Zero-Shot   & - &-& 65.4 & 95.8 &  - \\
Qwen2-VL-72B&Zero-Shot   & -&-& 64.9 & 94.7 &  - \\
\specialrule{\lightrulewidth}{0pt}{0pt}
\rowcolor[HTML]{FFF7F0}
\multicolumn{7}{c}{\rule[-1ex]{0pt}{3.3ex}\textbf{\textit{Student Models}}}\\
\specialrule{\lightrulewidth}{0pt}{0.8ex}
\multirow{3}{*}{InternVL2.5-4B}& Zero-Shot   & 8.8755 & 8.9423 &59.3 & 82.1  &  - \\
&LoRA & 3.4614 &3.8151& 62.3 & 87.8 &  3.7M  \\
&Full Params   & 3.0238& 3.2578& 63.4 & 90.7 & 3.0B\\
\midrule
\multirow{3}{*}{Qwen2.5-VL-3B}&Zero-Shot   & 8.0909 &8.1742& 61.2 & 85.8 &  - \\
&LoRA & 3.3101 &3.6461& 64.5 & 92.2 &  3.7M  \\

&Full Params   & 2.9127& 3.1017& 64.7 & 93.2 & 3.0B \\
\midrule
\multirow{3}{*}{Qwen2-VL-2B}& Zero-Shot   & 9.3210 &9.3872 & 58.7 & 80.4 &  - \\
&LoRA & 3.8127 & 4.3093& 61.3 & 86.8 & 4.4M   \\
&Full Params   &3.1846& 3.5204& 62.4 & 88.9 & 1.5B \\
\bottomrule
\end{tabular}
}
\label{tab:ab3}
\end{table*}


\subsection{Ablation Study}
\noindent \textbf{Component-level ablations.} Table~\ref{tab:ab1} validates the contributions of VideoStir's key components. Replacing the intent–relevance scorer with similarity-based frame retrieval results in substantial drops in both overall performance and retrieval accuracy, even when using strong vision–language embeddings from a powerful encoder such as PE. This indicates that pure semantic similarity matching can overlook the critical evidence required for long-video understanding, underscoring the importance of intent-aware retrieval. Meanwhile, removing the probability-weighted expectation (i.e., directly using discrete relevance levels) significantly degrades performance, suggesting that distributional scoring provides a smoother and better-calibrated ranking signal for discriminating frames. Moreover, disabling the spatio-temporal graph (i.e., flattening cues instead of performing structured retrieval), or removing spatial/temporal edges, markedly reduces retrieval accuracy, confirming that multi-hop aggregation over structured contexts is essential for extending evidence beyond the semantically matched anchor clips. In addition, replacing the event-boundary detector with uniform sampling that produces the same number of clips slightly degrades performance, suggesting that event-boundary segmentation isn’t the main performance driver, but it still improves evidence retrieval accuracy.

\noindent \textbf{Embedding backbones.} Table~\ref{tab:ab2} examines the impact of embedding choices on graph-based clip retrieval. Stronger video--language encoders yield consistent improvements on downstream performance. However, these gains are comparatively modest relative to the contributions from the framework’s core components, suggesting that VideoStir’s effectiveness is driven primarily by its structured design rather than by reliance on increasingly powerful embedding models.

\noindent \textbf{Scorer training strategy.} Table~\ref{tab:ab3} compares zero-shot inference, LoRA tuning, and full-parameter fine-tuning for the intent-relevance scorer. We evaluate different student-teacher pairs, including InternVL2.5~\cite{chen2024far}, Qwen2.5-VL~\cite{bai2025qwen2}, and Qwen2-VL~\cite{wang2024qwen2}. For Qwen2-VL-2B, we increase the LoRA rank to 32 (vs. 16 for others) to align trainable parameter counts for fair comparison. 

In the zero-shot setting, student's performance significantly behind teachers. Conversely, LoRA tuning markedly reduces validation cross-entropy on IR-600K and consistently improves LV-Bench performance, demonstrating robust generalization. Notably, LoRA captures the majority of full-param tuning gains using only about 4M parameters, in contrast to the billions required for full updates, with the downstream performance of LoRA-tuned Qwen2.5-VL-3B nearly matching the fully fine-tuned model. Consequently, we adopt the Qwen2.5-VL family for both teacher and student roles to maximize performance. These results confirm that IR-600K provides effective, transferable supervision, while LoRA offers an optimal accuracy–efficiency trade-off for practical long-video RAG systems.






\subsection{Discussion}
Tables~\ref{tab:comp1} and~\ref{tab:comp2} present a comprehensive comparison between VideoStir and SOTA baselines, including those that utilize external expert modules (e.g., captioners, object detectors, and OCR models) to augment video with auxiliary text. Our results indicate that VideoStir achieves superior performance using only video frames as native inputs, effectively bypassing the need for external textual augmentation. This suggests that one of the primary bottlenecks in long-video understanding still lies in the retrieval and organization of pivotal visual evidence. Therefore, refining the selection of visual evidence is essential for scaling MLLM's long-video understanding capabilities. VideoStir addresses this by enhancing both the relevance and spatio-temporal coherence of retrieved evidence, thereby facilitating more reliable reasoning in downstream MLLMs.


\section{Conclusion}
We propose VideoStir, a novel long-video RAG framework designed to address the limitations of contextual structure decoupling and intent misalignment prevalent in current paradigms. VideoStir models long videos as spatio-temporal graphs and utilizes multi-hop traversal to recover the intrinsic topology of their contexts, thereby aggregating spatio-temporally coherent evidence. Additionally, we introduce an intent-relevance scorer trained on our newly curated IR-600K dataset, which further filters visual content by capturing the query's underlying reasoning intent beyond surface-level semantics. Both the dataset and the scorer provide a solid foundation for future research into intent-oriented long-video RAG. Extensive experiments show that VideoStir is competitive with SOTA baselines without relying on auxiliary information, underscoring the effectiveness of shifting from flattened semantic matching to structured, intent-aware reasoning, and further revealing that one of the main bottlenecks in long-video understanding lies in retrieving and organizing pivotal visual evidence.

\nocite{fu2025brainvis,ren2025wamo,fu2025sdr,xiong2025thinking,ge2024can,fu2025vistawise,zhang2026test,liu2025structured,ti2025towards,wu2026camreasoner,ge2025mrfdmultiregionfusiondecoding}

\section*{Limitations}
As VideoStir introduces an additional step to transform long videos into structured representations, it inevitably incurs system latency. Indeed, system latency remains a broad challenge for long-video understanding frameworks that rely on external components like agents or RAG pipelines.
We regard further reducing end-to-end latency as a pivotal direction for future long-video RAG research.

\bibliographystyle{IEEEtran}
\bibliography{custom}

\newpage

\appendix

\section{The Use of LLMs}
The paper employs LLMs for language polishing.

\section{Extended discussion for Structured Retrieval}
\label{appendix:extended_case}
In Figure~\ref{fig:extended_case}, we further illustrate and discuss the advantages of structured retrieval via visualizations.
\begin{figure}[h]
    \centering
    \includegraphics[width=\linewidth]{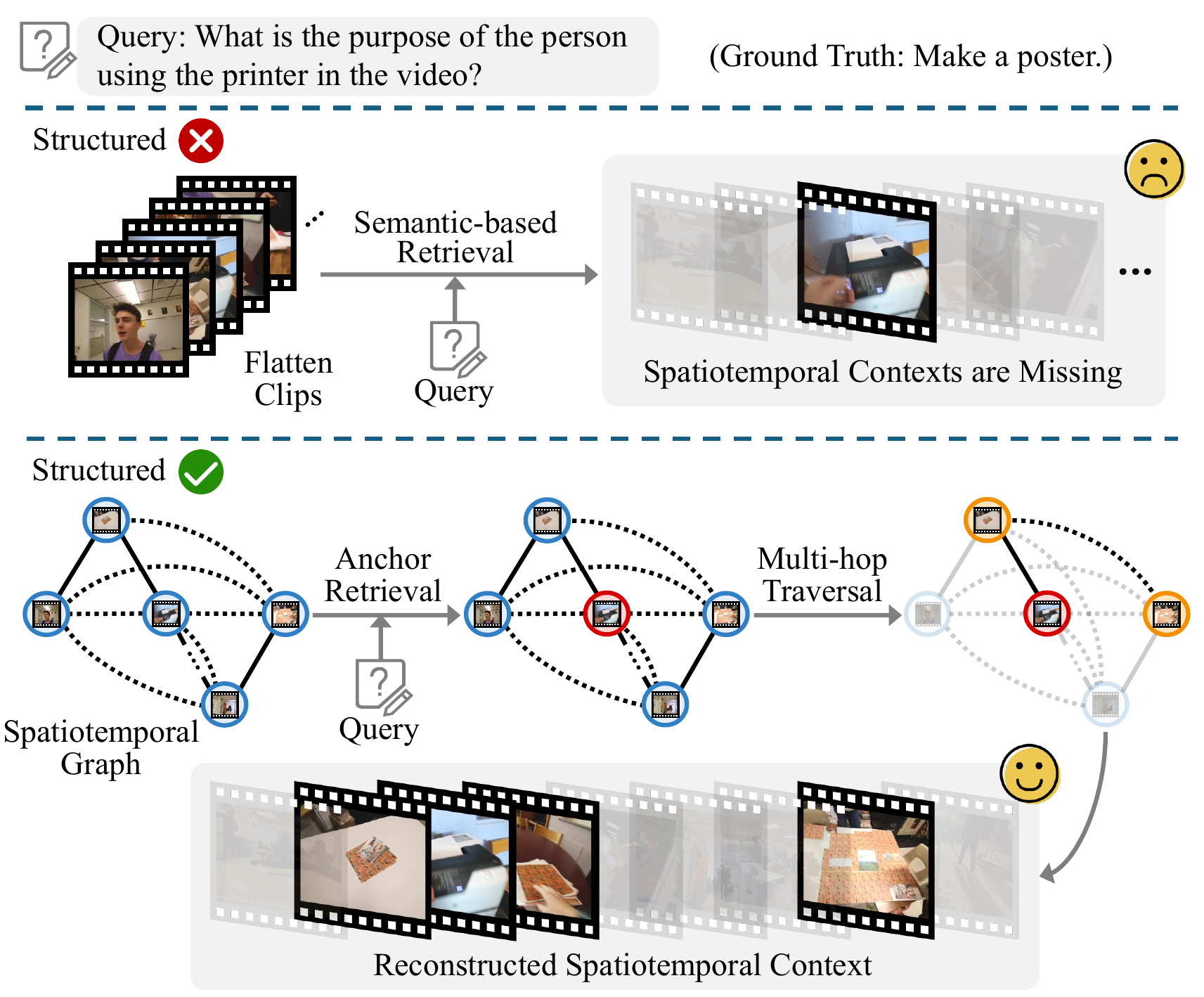}
    \caption{\small Advantages of structured retrieval over flat clip retrieval. Flattened retrieval mainly returns clips with explicit semantic overlap with the query, while overlooking clips that are contextually relevant but do not contain direct query-matching content. VideoStir structures long videos to reconstruct spatio-temporal context, allowing multi-hop traversal to aggregate contextually related clips around key moments. The resulting evidence is cleaner and more coherent, providing stronger support for subsequent fine-grained retrieval.}
    \label{fig:extended_case}
    \vspace{-0.1in}
\end{figure}

\section{Hyperparameter Ablations}
\label{appendix:hyperparam}
We determine the hyperparameters for the graph-based clip retrieval procedure based on the ablation study in Table~\ref{tab:hyperp}.

\begin{table}[h]
    \centering
    \caption{\small Hyperparameter ablations on Graph-based Clip Retrieval. We investigate the impact of the number of anchors $N$, hop distance $L$, and spatial edge threshold $\eta$. ``Avg. Clips'' denotes the average number of retrieved clips per query. The bold numbers represent the best results.}
    \label{tab:ablation_hyperparams}
    \resizebox{\linewidth}{!}{
    \begin{tabular}{lcccc}
        \toprule
        Hyperparameter & Value & Avg. Clips & Retrieval Acc. $\uparrow$ \\
        \midrule
        \multirow{4}{*}{\shortstack[l]{\textbf{Number of Anchors ($N$)} \\ \small \textit{(Fixed $L=2, \eta=0.6$)}}} 
          & 1 & 7.2 & 88.0  \\
          & \textbf{2} & 13.6 & \textbf{92.0}  \\
          & 3 & 21.9 & 90.5  \\
          & 4 & 25.4 & 90.0  \\
        \midrule
        \multirow{4}{*}{\shortstack[l]{\textbf{Hop Distance ($L$)} \\ \small \textit{(Fixed $N=2, \eta=0.6$)}}} 
          & 1 & 8.4 & 87.0  \\
          & \textbf{2}  & 13.6 & \textbf{92.0}  \\
          & 3 & 16.6 & 92.0 \\
          & 4 & 18.1 & 91.0  \\
        \midrule
        \multirow{4}{*}{\shortstack[l]{\textbf{Edge Weight Threshold ($\eta$)} \\ \small \textit{(Fixed $N=2, L=2$)}}} 
          & 0.2 & 31.4 & 84.0  \\
          & 0.4 & 20.6 & 90.5 \\
          & \textbf{0.6} & 13.6 & \textbf{92.0}  \\
          & 0.8 & 6.8 & 82.5  \\
        \bottomrule
    \end{tabular}
    }
    \label{tab:hyperp}
    \vspace{-0.1 in}
\end{table}

\section{Supplemental Details for IR-600K}
\label{appendix:dataset}

\subsection{Annotation Example}
We provide an example annotation below; during training, the text portion of the user content will be substituted into the intent-relevance scoring prompt as \{query\}.
\begin{tcolorbox}[
  fonttitle = \small\bfseries,
  title=Annotation Example,
  colframe=gray!2!black,
  colback=gray!2!white,
  boxrule=1pt,
  boxsep=0pt,
  left=5pt,
  right=5pt,
  fontupper=\setlength{\parskip}{6pt}\footnotesize, 
  halign title = flush center,
]

[\{"role": "system", "content": "You are a helpful assistant."\}, \{"role": "user", "content": [\{ "type": "image", "image": "8480330992/frame\_000310.jpg"\}, \{"type": "text", "text": "why do the ladies tap the children s heads with some soft item"\}]\}, \{ "role": "assistant", "content": [\{"type": "text","text": "2"\}]\}]

\end{tcolorbox}

\subsection{Dataset Distribution Statistics}
We summarize the data distributions of the IR-600K training and validation sets in Figures~\ref{fig:statis_train}. 

\begin{figure}[h]
    \centering
    \includegraphics[width=\linewidth]{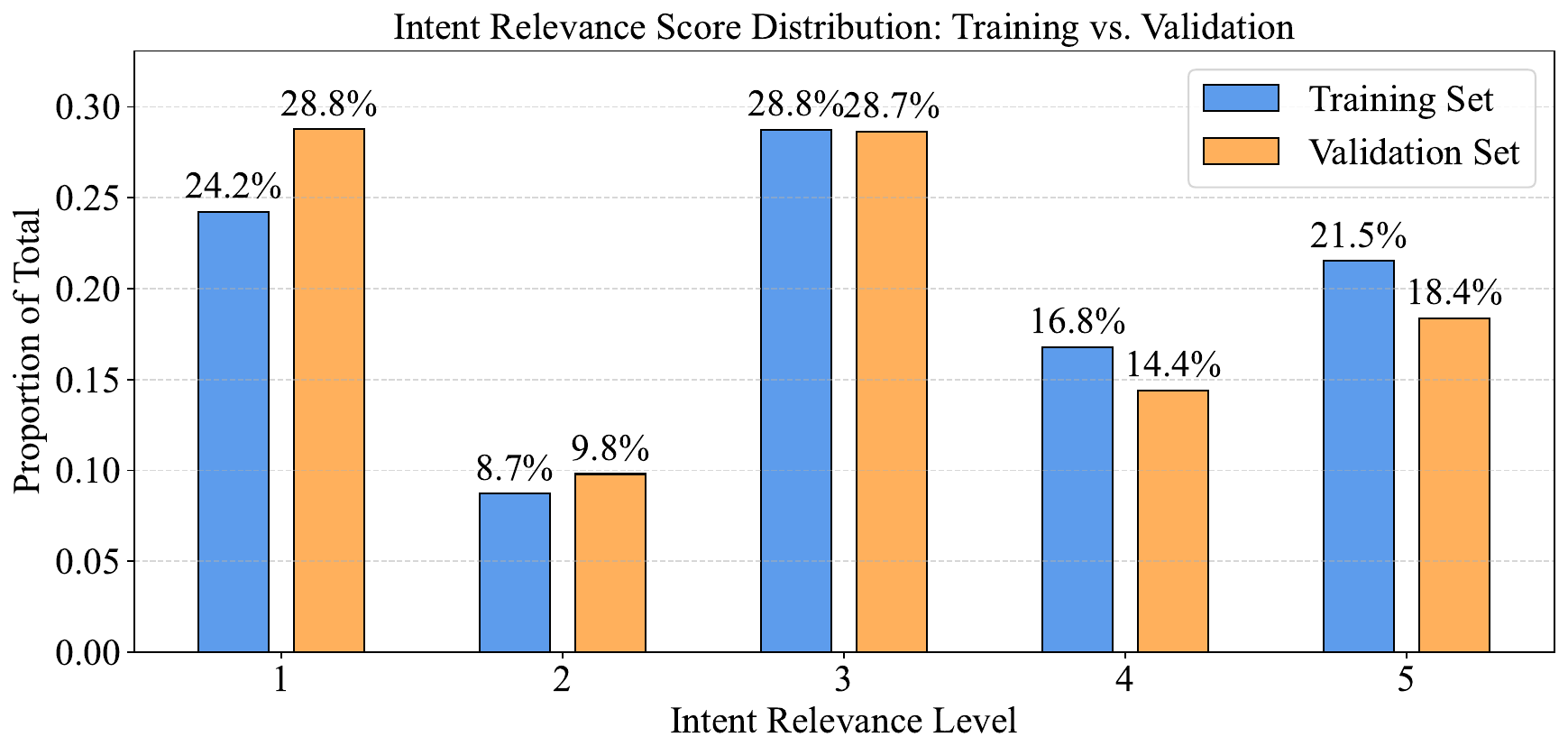}
    \caption{\small {Distribution of Intent Relevance Scores across the IR-600K training and validation sets. The results demonstrate that the validation set maintains a consistent class distribution with the training set across all relevance levels.}}
    \label{fig:statis_train}
\end{figure}

\section{Intent Relevance Scoring Prompt}
\label{appendix:prompt}

\begin{tcolorbox}[
  fonttitle = \small\bfseries,
  title=Intent-Relevance Scoring Prompt,
  colframe=gray!2!black,
  colback=gray!2!white,
  boxrule=1pt,
  boxsep=0pt,
  left=5pt,
  right=5pt,
  fontupper=\setlength{\parskip}{6pt}\footnotesize, 
  halign title = flush center,
]

Infer the query’s intent and evaluate how likely it is that this frame is intent-relevant for answering: ‘\{query\}’. 

Output only one number from 1 to 5, where:

[1] = completely irrelevant — the frame provides no visual or contextual information related to the question or its answer.

[2] = slightly relevant — the frame shows general background or context, but it is unlikely to contribute to answering.

[3] = moderately relevant — the frame includes partial clues or indirect context that might help infer the answer, but the key evidence is missing.

[4] = mostly relevant — the frame provides substantial visual or contextual information that can be used to answer the question, though not fully decisive.

[5] = highly relevant — the frame clearly contains the decisive evidence or strong contextual cues that directly or indirectly support the correct answer.

\end{tcolorbox}

\end{document}